\documentclass[12pt]{article}
\usepackage{amsfonts}
\usepackage{amssymb}
\usepackage{fullpage}
\usepackage{amsmath}
\usepackage{rotating}
\usepackage{natbib}
\usepackage[doublespacing, nodisplayskipstretch]{setspace}
\usepackage{titlesec}
\usepackage{makecell}

\usepackage{tikz}

\usepackage{xcolor}
\usetikzlibrary{decorations.pathreplacing}
\usepackage[margin=1in, footskip=0.4in]{geometry}

\usepackage{graphicx}
\graphicspath{{../Code/outputs/}}

\usepackage[font=small,labelfont=bf]{caption}
\usepackage{subcaption}

\usepackage[colorlinks=true, citecolor=blue]{hyperref}

\begin{document}
	

\title{Zero is Not Hero Yet: Benchmarking Zero-Shot Performance of LLMs for Financial Tasks\footnote{This working paper is an ongoing research project, and feedback is greatly appreciated. }}

\author{ {\large Agam Shah\footnote{Center for Machine Learning, School of Computational Science \& Engineering, Georgia Institute of Technology; \textit{Email:} \texttt{ashah482@gatech.edu}}}  
		\and {\large Sudheer Chava\footnote{Scheller College of Business at Georgia Institute of Technology; \textit{Email:} \texttt{chava@gatech.edu}}} }
\date{This Version: \today}


\maketitle
\thispagestyle{empty}

\doublespacing
\normalsize

\vspace{24pt}
\begin{abstract}
\noindent Recently large language models (LLMs) like ChatGPT have shown impressive performance on many natural language processing tasks with zero-shot. In this paper, we investigate the effectiveness of zero-shot LLMs in the financial domain. We compare the performance of ChatGPT along with some open-source generative LLMs in zero-shot mode with RoBERTa fine-tuned on annotated data. We address three inter-related research questions on data annotation, performance gaps, and the feasibility of employing generative models in the finance domain. Our findings demonstrate that ChatGPT performs well even without labeled data but fine-tuned models generally outperform it. Our research also highlights how annotating with generative models can be time-intensive. Our codebase is publicly available on GitHub under CC BY-NC 4.0 license\footnote{The code is available on \href{https://github.com/gtfintechlab/zero-shot-finance}{FinTech Lab GitHub}.}. 

\end{abstract}

\clearpage
\setcounter{page}{1}

\section{Introduction}
On November 30th, 2022, OpenAI released ChatGPT\footnote{\url{https://chat.openai.com/}} and intrigued the world with its capabilities. Within a few months after its release, researchers have started testing its zero-shot capabilities for financial domain tasks. \cite{shah2023trillion} and \cite{hansen2023can} have demonstrated the use of ChatGPT to decode the communications from the Federal Open Market Committee (FOMC) and how it can be used in understanding financial markets. \citet{saggu2023influence} present evidence that ChatGPT is already influencing AI-related crypto assets. Despite the impressive demonstrations using prompts, it is still important to further understand ChatGPT capabilities across various NLP tasks in the financial domain. 

\citet{pikuliak_chatgpt_survey} conducted a survey to understand the linguistic capabilities of ChatGPT in solving natural language processing (NLP) tasks and found that ChatGPT outperforms fine-tuned models only on 22.5\% (34 out of 151) tasks. \citet{qin2023chatgpt} empirically analyzes the zero-shot learning ability of ChatGPT across 20 NLP datasets, highlighting its effectiveness (e.g., arithmetic reasoning) and limitations (e.g., sequence tagging), and provides qualitative case studies for further analysis. Even though researchers have done work to understand the capabilities of ChatGPT for general domain NLP tasks, there is no work done for financial NLP. To fill in this gap, we benchmark the zero-shot performance of ChatGPT and compare it with fine-tuned RoBERTa for various financial NLP tasks. Compared to previous work where they report fine-tuning numbers from existing work, we run all experiments ourselves. 

The recent op-ed article by \citet{rogers-etal-2023-closed} argues that closed AI models like ChatGPT make bad baselines and make a case that these models shouldn't be a requisite baseline in scientific work. To provide an open-source alternative to ChatGPT, some organizations have released competing open-source LLMs. Databricks fine-tuned pythia-12b \citep{pythia-12b} model on approximately 15k instruction fine-tuning records generated by Databricks employees and released dolly-v2-12b \citep{dolly-v2-12b}. Following the release of dolly-v2-12b, H2O.ai released h2ogpt-oasst1-512-12b \citep{h2ogpt-oasst1-512-12b} which is developed by fine-tuning pythia-12b on open-source instruct-type dataset \citep{openassistant_oasst1_h2ogpt_graded}. In order to understand where these open-source models stand in comparison with ChatGPT, we also benchmark these models for all the tasks in our study. To also understand how good these instruction following models are in following instructions, we also report \% of test samples where they don't follow instructions. 

In the last few years, language models have scaled exponentially. BERT-large has \citep{devlin2018bert} 345 million parameters vs LLaMA \citep{touvron2023llama} has 65 billion parameters, GPT-3 \citep{brown2020language} has 175 billion parameters and GPT-4 is estimated/rumored to have a few trillion parameters. For a comprehensive review of LLMs, we direct the reader's attention to the extensive surveys conducted by \citet{mialon2023augmented} and \citet{zhao2023survey}. Given that latency is of high importance to many financial applications, we not only compare the performance of models but also compare the time it takes to label one sentence. It also helps us understand how feasible it is to use for a research project which requires labeling large datasets. 

In summary through this work, we try to answer the following research questions: 
\begin{itemize}
    \item \textbf{RQ1: } For financial domain tasks, is it better to annotate the data and fine-tune a medium-sized model instead of using a generative LLM with zero-shot prompt? 
     \item \textbf{RQ2: } What is the performance gap between ChatGPT-like closed model and open-sources LLMs for financial domain tasks? 
    \item \textbf{RQ3: } Is it feasible to employ generative LLMs for research when the text data is large? 
\end{itemize}

In order to answer these questions, we use four financial NLP tasks in our study and benchmark various models on all of them. We employ RoBERTa-base and RoBERTa-large models for fine-tuning benchmarks while using ChatGPT-3.5-Turbo, Dolly-V2-12B, and H2O-12B as zero-shot models. 

\paragraph{Key insights} To the best of our knowledge this is the first study that not only attempts to understand how well ChatGPT can perform with zero-shot on many nlp tasks but compares it with other open-source generative LLMs and fine-tuned PLMs\footnote{Throughout the paper we refer to the earlier models built based on BERT architecture as PLMs and latest generative models similar to GPT models as LLMs. } for the financial domain. The key takeaways are summarized as follows: 
\begin{itemize}
    \item Even though zero-shot ChatGPT fails to outperform fine-tuned PLMs, it provides impressive performance across all the tasks without having access to any labeled data. 
    \item The performance gap between fine-tuned model PLMs and ChatGPT is larger where the dataset is not publicly available yet. It will be an interesting future study to understand plausible contamination issues as a potential explanation. 
    \item The performance of fully open-source LLMs for all financial tasks is significantly lower as compared to ChatGPT. 
    \item In certain scenarios, even if the user is willing to accept the performance difference between zero-shot LLMs and fine-tuned PLMs, the amount of time required to assign labels to data is 1000 times greater when using generative LLMs.
\end{itemize}

\section{Datasets and Tasks}
We include hawkish-dovish sequence classification from \citet{shah2023trillion}, financial sentiment analysis task from \citet{malo2014good}, financial numerical claim detection from \citet{shah2022numerical}, and named entity recognition dataset from \citet{shah2023finer}. A summary of datasets used with the train-validation-test split is provided in table \ref{tb:dataset_summary}.

\begin{table*}[ht]
\centering

\begin{tabular}{llccc}
\hline
Task            & Source      & \multicolumn{3}{c}{Dataset Size}  \\
                &             & Train & Valid & Test \\

\hline
Hawkish-Dovish-Neutral Classification & \citet{shah2023trillion} & 1587 & 397 & 496\\
Sentiment Classification  & \cite{malo2014good} & 1449 & 362 & 453\\
Claim Detection           & \citet{shah2022numerical} & 1715 & 429 & 537\\
Named Entity Recognition  & \citet{shah2023finer} & 80.5k & 10.2k & 26k \\
\hline
\end{tabular}

\caption{Summary of benchmarks used. Dataset size denotes the number of samples in the benchmark. For the NER task each sample consists of a word token while for other tasks it represents a sentence token.}
\label{tb:dataset_summary}
\end{table*}

\subsection{FOMC Communication}
Understanding or deciphering the monetary policy stance of the central banks is an important NLP task to understand financial markets better. Several studies \citep{rozkrut2007quest, tobback2017between, hansen2018transparency, cieslak2019stock, tsukioka2020tone, bennani2020does, shah2023trillion} have found that the words used by central banks have an impact on the market, although the extent of this impact varies depending on the communication style and chair. Recent work by \citet{hansen2023can} uses ChatGPT to decode the FOMC post-meeting statements. In order to understand the capabilities of LLMs in decoding central bank communications, we use open-source labeled data that combines meeting minutes, press conference transcripts, and speeches \citep{shah2023trillion}. In this dataset, each sentence is labeled one of the three (hawkish, dovish, and neutral) labels which we use as a sequence classification task. 

\subsection{Sentiment Analysis}
Sentiment analysis is a popular task in the financial domain as it correlates with the market sentiment which influences price movements. In the finance literature, the sentiment dictionary developed by \citet{loughran2011liability} is used as bag-of-words extensively for sentiment analysis. \cite{garcia2023colour} refine the dictionary for bag-of-words and claim it to be SOTA in the finance literature. On the other side computational linguistics literature has grown exponentially over the last decade. After the introduction of sentence-level financial sentiment data by \citet{malo2014good}, many models have been trained for sentiment analysis tasks. Soon after the release of BERT \citep{devlin2018bert}, FinBERT \citep{araci2019finbert} was developed for financial sentiment analysis. To understand how well generative LLMs can do on this important task, we use Financial Phrasebank sentiment analysis data\footnote{\url{https://huggingface.co/datasets/financial_phrasebank}} developed by \citet{malo2014good} for the financial sentiment classification task. The dataset contains multiple versions. Here we use the data where there is 100\% annotation agreement. We perform three class (positive, negative, and neutral) sequence classification task. 

\subsection{Numerical Claim Detection}
Extraction of numerical claims from the financial text like analysts' reports, earnings calls, news, etc is helpful in forecasting volatility of the stock prices. For the task of claim detection, we use a dataset developed by \citet{shah2022numerical}. The dataset contains binary labels (``in-claim", and ``out-of-claim") for sentences extracted from a heterogenous set of analysts' reports. In this case, the term ``in-claim" text is used to refer to sentences in the financial domain that contain specific and measurable financial claims, which are not factual statements. For instance, the sentence ``Operating income is expected to be between \$2.1 billion and \$3.6 billion" is considered an "in-claim" sentence because it predicts a future outcome. On the other hand, the sentence ``Revenues increased by 48.6\% compared to the previous year, reaching \$5.44 billion, primarily due to the expansion of the customer base" is categorized as ``out-of-claim" since it presents factual information from the past.

\subsection{Named Entity Recognition}
Named Entity Recognition (NER) involves the identification and classification of named entities in text, such as person names, organization names, and locations, among others. Given a sentence with N tokens and a set of entities denoted as S with a size of \#S, the NER model assigns an entity label to each token. Following the BIO convention, the label set L has a size of $2\#S + 1$ and includes "O" to indicate categories not listed in the entity list. NER plays a crucial role in financial NLP due to its significant implications for various applications in this domain. NER enables the extraction and categorization of key financial entities such as company names, person names, locations, etc. This facilitates the identification of important entities involved in financial news and reports, helping analysts and investors gain insights into market trends, company performance, and potential investment opportunities. For instance, NER can aid in tracking the mentions of specific companies in news articles, social media posts, and financial reports, enabling the monitoring of market sentiment and the detection of emerging patterns that may impact stock prices. Therefore, the accurate identification and classification of named entities through NER are crucial for extracting meaningful information and enhancing decision-making processes in financial NLP. For financial NER we use a dataset developed in \citet{shah2023finer}.

\section{Experiments}
We run all the experiments for this paper and do not report numbers from any prior work. We use 3 different seeds to split datasets into train and test parts except for FiNER-ORD as it has a train-validation-test split already given. 

\subsection{Fine-Tuning PLM}
In order to set the benchmark, we use base (``roberta-base") and large (``roberta-large") versions of RoBERTa \citep{liu2019roberta} model. No pre-training is conducted on the models before proceeding with the fine-tuning process. To determine the optimal hyper-parameters for each model, a grid search is performed using four different learning rates (1e-4, 1e-5, 1e-6, 1e-7) and four different batch sizes (32, 16, 8, 4). We use a maximum of 100 epochs for training with early stopping criteria. If the validation F1 score doesn't improve by more than or equal to 1e-2 in the next 7 epochs then we use the best model stored earlier as the final fine-tuned model. The experiments are carried out using PyTorch \citep{pytorch} on an NVIDIA RTX A6000 GPU. The initialization of each model is based on the pre-trained version available in the Huggingface Transformers library \citep{huggingface}. 

\subsection{Zero-Shot with Generative LLMs}
In the generative LLM category, we utilize ChatGPT ("gpt-3.5-turbo") with specific settings including a maximum token limit of 1000 and a temperature value of 0.0. As for the open-source LLM category, we employ "dolly-v2-12b" and "h2ogpt-oasst1-512-12b" along with their dedicated text generation pipelines and models, which can be found on their respective Huggingface pages.

For each task, we design separate prompts and write functions that can be used to get labels from the output of the prompt. As zero-shot learning doesn't require any labeled data for training, we only use test split for prompting. The prompt template for each task is discussed below.

\paragraph{FOMC Communication}
We use the following zero-shot prompt for hawkish-dovish-neutral classification: 

``Discard all the previous instructions. Behave like you are an expert sentence classifier. Classify the following sentence from FOMC into `HAWKISH', `DOVISH', or `NEUTRAL' class. Label `HAWKISH' if it is corresponding to tightening of the monetary policy, `DOVISH' if it is corresponding to easing of the monetary policy, or `NEUTRAL' if the stance is neutral. Provide the label in the first line and provide a short explanation in the second line. The sentence: \{sentence\}"

\paragraph{Sentiment Analysis}
We use the following zero-shot prompt for sentiment classification: 

``Discard all the previous instructions. Behave like you are an expert sentence sentiment classifier. Classify the following sentence into `NEGATIVE', `POSITIVE', or `NEUTRAL' class. Label `NEGATIVE' if it is corresponding to negative sentiment, `POSITIVE' if it is corresponding to positive sentiment, or `NEUTRAL' if the sentiment is neutral. Provide the label in the first line and provide a short explanation in the second line. The sentence:  \{sentence\}"

\paragraph{Numerical Claim Detection}
We use the following zero-shot prompt for numerical claim detection: 

``Discard all the previous instructions. Behave like you are an expert sentence sentiment classifier. Classify the following sentence into `INCLAIM', or `OUTOFCLAIM' class. Label `INCLAIM' if consist of a claim and not just factual past or present information, or `OUTOFCLAIM' if it has just factual past or present information. Provide the label in the first line and provide a short explanation in the second line. The sentence:  \{sentence\}"

\paragraph{Named Entity Recognition}
We use the following zero-shot prompt for named entity recognition: 

``Discard all the previous instructions. Behave like you are an expert named entity identifier. Below a sentence is tokenized and each line contains a word token from the sentence. Identify `Person', `Location', and `Organisation' from them and label them. If the entity is multi token use post-fix \_B for the first label and \_I for the remaining token labels for that particular entity. The start of the separate entity should always use \_B post-fix for the label. If the token doesn't fit in any of those three categories or is not a named entity label it `Other'. Do not combine words yourself. Use a colon to separate token and label. So the format should be token:label. \textbackslash n\textbackslash n \{word tokens separated by \textbackslash n\}"

\section{Results}
In this section, we benchmark and evaluate all the models and tasks discussed in previous sections. For performance analysis, we report the mean and standard deviation of the weighted F1 score. Along with performance for each model, we also report the time it takes on average for each model to label a sentence in the test dataset. For fine-tuned PLMs, we additionally report fine-tuning time for the best hyperparameter setting. For generative LLMs, in many cases, the model doesn't follow the instruction. We assign the label ``-1" (which makes their weight 0 while calculating the weighted F1 score) to those instances and report percentage of those instances as ``Missing \%". 

\subsection{FOMC Communication}
Performance and other metrics for the FOMC tone classification task are reported in table \ref{tb:fomc_communication_results}. Fine-tuned RoBERTa-base and RoBERTa-large have similar performance and they outperform zero-shot LLMs. In the zero-shot LLM category, ChatGPT outperforms other open-source LLMs by a good margin. ChatGPT also follows the instruction 100\% of the time, while Dolly fails to follow instructions for 6.05\% of the time and H2O for 42.14\% of the time. We run all the codes before we made the data for this task public as part of our other work \citep{shah2023trillion} to ensure that there is no contamination issue. The results for fine-tuning models are a little different than those reported in \citet{shah2023trillion} as we employ different early stopping mechanism compared to them. Even if for a particular use case performance is not an issue, the time it takes for generative LLM to annotate the data is 1000 times higher compared to fine-tuned PLMs. The latency of a few seconds is not useful when markets might adjust the price in a few milliseconds. 

\begin{table*}[ht]
\centering
\footnotesize
\begin{tabular}{lcccc}
\hline
\textbf{Model} & \textbf{Fine Tuning Time} & \textbf{Test Labeling Time} & \textbf{F1 Score mean (std)} & \textbf{Missing \%} \\
\hline
\multicolumn{5}{c}{Panel A: Fine-Tuning with PLM}\\
\hline
RoBERTa-base & 5.85 minutes & 3.85 milliseconds & 0.6990 (0.0182) & - \\
RoBERTa-large & 18.11 minutes & 9.87 milliseconds & 0.6977 (0.0110) & - \\
\hline
\multicolumn{5}{c}{Panel B: Zero-Shot with Generative LLM}\\
\hline
ChatGPT-3.5-Turbo & - & **3.91 seconds & 0.5837 (0.0155) & 0.00\% \\
Dolly-V2-12B & - & 5.89 seconds & 0.1195 (0.0100) & 6.05\% \\
H2O-12B & - & 29.64 seconds & 0.0915 (0.0105) & 42.14\% \\
\hline
\end{tabular}
\caption{Experiment results on hawkish-dovish-neutral classification task. An average of 3 seeds was used for all models. **It is based on an API call, for more details check the ``Limitations and Future Work" section.}
\label{tb:fomc_communication_results}
\end{table*}

\subsection{Sentiment Analysis}
For the sentiment analysis task, performance and other metrics are reported in table \ref{tb:fpb_sentiment_results}. The results follow a similar trend but for this task, the H2O model doesn't follow the instruction at all which is surprising. It will be an interesting future study to understand why this is the case and how open-source LLMs can be improved in this dimension. The ChatGPT achieves impressive performance close to 0.9 F1 score for sentiment analysis which is not far from the performance of fine-tuned RoBERTa.

\begin{table*}[ht]
\centering
\footnotesize
\begin{tabular}{lcccc}
\hline
\textbf{Model} & \textbf{Fine Tuning Time} & \textbf{Test Labeling Time} & \textbf{F1 Score mean (std)} & \textbf{Missing \%} \\
\hline
\multicolumn{5}{c}{Panel A: Fine-Tuning with PLM}\\
\hline
RoBERTa-base  & 4.89 minutes & 2.62 milliseconds & 0.9735 (0.0041) & - \\
RoBERTa-large & 6.50 minutes & 4.50 milliseconds & 0.9757 (0.0077) & - \\
\hline
\multicolumn{5}{c}{Panel B: Zero-Shot with Generative LLM}\\
\hline
ChatGPT-3.5-Turbo & - & **4.11 seconds & 0.8929 (0.0078) & 0.00\% \\
Dolly-V2-12B & - & 6.93 seconds & 0.1070 (0.0132) & 3.16\% \\
H2O-12B & - & 54.66 seconds & 0.0000 (0.0000) & 100.00\% \\
\hline
\end{tabular}
\caption{Experiment results on the sentiment classification task. An average of 3 seeds was used for all models. **It is based on an API call, for more details check the ``Limitations and Future Work" section.}
\label{tb:fpb_sentiment_results}
\end{table*}

\subsection{Numerical Claim Detection}
Performance and other metrics for the numerical claim detection task are reported in table \ref{tb:numerical_claim}. The gap between fine-tuned model PLMs and ChatGPT is larger here compared to the sentiment analysis task while the gap between Dolly and ChatGPT is smaller. The exact reason for this is uncertain, but one possibility could be contamination, considering that the sentiment analysis dataset is publicly available while the claim dataset is not. Exploring the potential contamination issues as an explanation for this discrepancy would be an intriguing avenue for future research. Other results follow a similar trend. 

\begin{table*}[ht]
\centering
\footnotesize
\begin{tabular}{lcccc}
\hline
\textbf{Model} & \textbf{Fine Tuning Time} & \textbf{Test Labeling Time} & \textbf{F1 Score mean (std)} & \textbf{Missing \%} \\
\hline
\multicolumn{5}{c}{Panel A: Fine-Tuning with PLM}\\
\hline
RoBERTa-base  & 2.44 minutes & 1.60 milliseconds & 0.9615 (0.0091) & - \\
RoBERTa-large & 10.41 minutes & 6.31 milliseconds & 0.9642 (0.0069) & - \\
\hline
\multicolumn{5}{c}{Panel B: Zero-Shot with Generative LLM}\\
\hline
ChatGPT-3.5-Turbo & - & **5.51 seconds & 0.8136 (0.0079) & 0.00\% \\
Dolly-V2-12B & - & 5.66 seconds & 0.5250 (0.0203) & 5.59\% \\
H2O-12B & - & 54.97 seconds & 0.0536 (0.0054) & 42.89\% \\
\hline
\end{tabular}
\caption{Experiment results on the numerical claim detection task. An average of 3 seeds was used for all models. **It is based on an API call, for more details check the ``Limitations and Future Work" section.}
\label{tb:numerical_claim}
\end{table*}

\subsection{Named Entity Recognition}
Table \ref{tb:ner_finer_ord} presents the performance metrics for the Named Entity Recognition (NER) task. Similar to the previous tasks, the results exhibit a consistent pattern. However, the NER prompt proves to be more challenging compared to other tasks, leading both open-source models to struggle in adhering to the given instruction. Moreover, decoding labels from the prompt output is relatively more difficult in token classification compared to sequence classification tasks, resulting in a higher percentage of samples with missing labels for generative LLMs in the NER task. 

\begin{table*}[ht]
\centering
\footnotesize
\begin{tabular}{lcccc}
\hline
\textbf{Model} & \textbf{Fine Tuning Time} & \textbf{Test Labeling Time} & \textbf{F1 Score mean (std)} & \textbf{Missing \%} \\
\hline
\multicolumn{5}{c}{Panel A: Fine-Tuning with PLM}\\
\hline
RoBERTa-base  & 14.22 minutes & 3.37 milliseconds & 0.9696 (0.0060) & - \\
RoBERTa-large & 29.04 minutes & 10.57 milliseconds & 0.9754 (0.0047) & - \\
\hline
\multicolumn{5}{c}{Panel B: Zero-Shot with Generative LLM}\\
\hline
ChatGPT-3.5-Turbo & - & **9.49 seconds & 0.8509 & 17.59\% \\
Dolly-V2-12B & - & 6.59 seconds & 0.0023 & 99.88\% \\
H2O-12B & - & 44.82 seconds & 0.0056 & 99.67\% \\
\hline
\end{tabular}
\caption{Experiment results on the named entity recognition task. An average of 3 seeds was used for fine-tuned PLMs. For generative LLMs, we only have one test split so the standard deviation is not reported. **It is based on an API call, for more details check the ``Limitations and Future Work" section.}
\label{tb:ner_finer_ord}
\end{table*}

\section{Conclusion}
\label{sec:conclusion}
In conclusion, we investigate the effectiveness of using ChatGPT, in zero-shot mode as compared to other open-source generative LLMs and fine-tuned PLMs in the financial domain. The key findings and implications derived from the research are as follows:

Firstly, regarding RQ1, it was observed that while fine-tuned PLMs generally outperformed zero-shot ChatGPT, the latter still demonstrated impressive performance across various tasks without the need for labeled data. This indicates that ChatGPT has the potential to be a ``hero" for financial domain tasks even without explicit fine-tuning.

Secondly, in relation to RQ2, a notable performance gap was identified between fine-tuned PLMs and ChatGPT, particularly in cases where the dataset was not publicly available. This discrepancy could be attributed to potential contamination issues within the data. 

Additionally, concerning RQ3, we examine the feasibility of employing generative LLMs for research purposes when dealing with large volumes of textual data requiring annotation. Our findings indicate that despite potential performance gaps between zero-shot LLMs and fine-tuned PLMs, the time required to label a single sample using generative LLMs was significantly higher, potentially by a factor of 1000. 

Overall, to our knowledge, this is the first paper to evaluate the performance of ChatGPT in zero-shot mode across multiple NLP tasks and compare it with other open-source LLMs and fine-tuned PLMs in the financial domain. Our results emphasize the potential and limitations of ChatGPT, highlighting the trade-offs between performance, availability of labeled data, and labeling efficiency. Future research in this area could explore strategies to mitigate contamination issues in closed models and address the labeling time challenges posed by generative LLMs.

\section*{Limitations and Future Work}
\label{sec:limitations}
The range of tasks used here is not exhaustive and we plan to keep adding more tasks in the future from Financial Language Understanding
Evaluation (FLUE) \citep{shah2022flue} and other sources. In future drafts, we also plan to add GPT-4, Alpaca, MPT, and other models. In this study, we do not benchmark few-shot as the focus is to understand how far zero-shot is from fine-tuned models. We also do not include and discuss finance domain-specific LLMs like BloombergGPT \citep{wu2023bloomberggpt} in our work as we have no way of accessing it. 

For all the experiments regarding ChatGPT, we use API calls which limits our ability to accurately compare the time it takes for it takes to label a sentence. For all other models, we use the same hardware with NVIDIA RTX A6000 48GB GPU and we don't know what hardware or GPU OpenAI uses. We also note that the study neither tries to understand biases nor looks at the contamination issues with LLMs. It attempts to ask questions that ``if there is no contamination issue, where do current open and closed LLMs stand compared to fine-tuned models when it comes to solving a task in the financial domain?". 

\section*{Acknowledgements}
We appreciate the generous infrastructure support provided by Georgia Tech's Office of Information Technology, especially Robert Griffin. We would also like to thank Agoston Reguly, and Arnav Hiray for their comments and suggestions. 

\singlespacing
\newpage
\bibliographystyle{acl_natbib}
\bibliography{references}

\end{document}